\documentclass{llncs}

\usepackage[utf8]{inputenc}
\usepackage{amsmath}
\usepackage{todonotes}
\usepackage{xspace}
\usepackage{hyperref}

\usepackage{colortbl}
\usepackage{booktabs}
\usepackage{float}
\usepackage{array}
\definecolor{cellblack}{RGB}{169, 169, 169}
\newcommand{\cell}[1]{\vspace{-0.83em}\color[hsb]{#1,1,1}\rule[-3.4pt]{2em}{2em}}
\newcommand{\emptycell}[0]{\vspace{-0.83em}\color{white}\rule[-3.2pt]{2em}{2em}}
\newcommand{\probcell}[2]{\centering\fcolorbox[hsb]{#1,1,1}{#1,1,1}{#2\strut}}

\newcolumntype{S}{@{}
     p{2em} @{}
}

\newcommand{\turnstile}{\hspace{1mm}:\hspace{-1mm}-\hspace{1mm}}
\usepackage{subcaption}
\title{ProbLife: a Probabilistic Game of Life}
\author{Simon Vandevelde\thanks{This research received funding from the Flemish Government under the “Onderzoeksprogramma Artificiële Intelligentie (AI) Vlaanderen” programme.} \and Joost Vennekens}
\institute{KU Leuven, De Nayer Campus, Dept. of Computer Science \\
Leuven.AI - KU Leuven Institute for AI, B-3000 Leuven, Belgium \\ 
\email{\{s.vandevelde, joost.vennekens\}@kuleuven.be}}  

\begin{document}

\maketitle
\begin{abstract}
    This paper presents a probabilistic extension of the well-known cellular automaton, Game of Life.
    In Game of Life, cells are placed in a grid and then watched as they evolve throughout subsequent generations, as dictated by the rules of the game.
    In our extension, called ProbLife, these rules now have probabilities associated with them.
    Instead of cells being either dead or alive, they are denoted by their chance to live.
    After presenting the rules of ProbLife and its underlying characteristics, we show a concrete implementation in ProbLog, a probabilistic logic programming system.
    We use this to generate different images, as a form of rule-based generative art.
\end{abstract}

\section{Introduction}

Game of Life (or Life) \cite{gol} is a well-known cellular automaton invented by John Conway, which takes place in a rectangular grid consisting of cells that are either ``dead'' or ``alive''.
The grid goes through multiple generations, simulating ``evolution'', in which cells can die, survive, or be born based on their number of living neighbours.
It is often called a 0-player game: after selecting an initial state of cells, we sit back and watch the grid evolve through time.

In this paper we present ProbLife, which extends Game of Life with a probability element and continuous cell values in the range of $[0..1]$.
While there already exist many extensions and variations of Life (and other cellular automata) that include probabilistic elements, each of these introduces the concept of ``probability'' in different ways.
ProbLife distinguishes itself in two aspects: (a) rules in ProbLife have a probability associated to them and (b) instead of being limited to binary values, the cells can have any value in the continuous range of $[0..1]$.

To play ProbLife, we create a practical implementation in the form of a probabilistic logic program in ProbLog.
This approach allows us to elegantly represent the logic of ProbLife, in a flexible manner.
As such, it can be used to quickly prototype different rulesets and experiment with the associated probabilities.

The act of generating grids from an initial state based on a predefined set of rules can be classified as a form of rule-based generative art \cite{boden2009}.
While the grids generated by standard Game of Life can only be visualised dichromatically (typically in black and white), the cells in ProbLife, due to their continuous nature, can be drawn using colour gradients.

In short, the contributions of this paper are:
\begin{itemize}
    \item an overview of the Game of Life variants that include probabilities;
    \item the presentation of ProbLife, a probabilistic extension of Game of Life;
    \item a concrete implementation of ProbLife in ProbLog.
\end{itemize}

This paper is structured as follows.
In Section \ref{s:gol}, we elaborate on the specifics of Life and its variants, with a specific focus on those with probabilistic or continuous elements.
Afterwards, we present ProbLife and its rules in Section \ref{s:problife}, and show a concrete ProbLog implementation in Section \ref{s:problog}.
Finally, in Section \ref{s:discussion} we discuss ProbLife in relation to the other probabilistic variants, present some interesting ProbLife instances we were able to find, and conclude.

\section{Game of Life, extensions and variants}\label{s:gol}

\begin{figure}
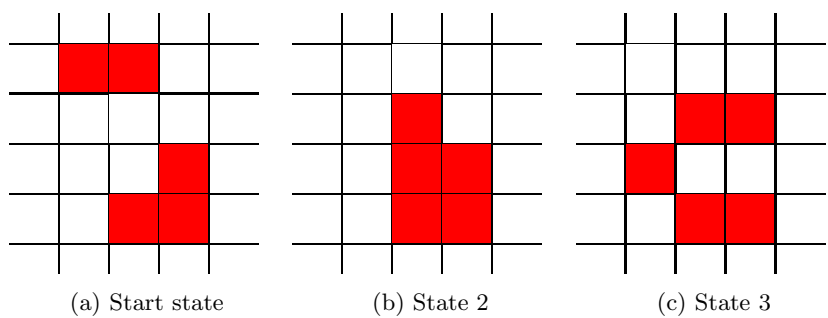

    \centering
    \begin{subfigure}[b]{0.3\textwidth}
        \begin{tabular}{ S | S | S | S | S }
            & & & & \\
            \hline
            & \cell{0.0} & \cell{0.0} & &\\
            \hline
            & & \emptycell & & \\
            \hline
            & & & \cell{0.0} &  \\
            \hline
            & & \cell{0.0}& \cell{0.0} & \\
            \hline
            & & & & \\
        \end{tabular}
        \caption{Start state}
    \end{subfigure}
    \begin{subfigure}[b]{0.3\textwidth}
        \begin{tabular}{ S | S | S | S | S }
            & & & & \\
            \hline
            & & \emptycell & &\\
            \hline
            & & \cell{0.0} & & \\
            \hline
            & & \cell{0.0} & \cell{0.0} &  \\
            \hline
            & & \cell{0.0}& \cell{0.0} & \\
            \hline
            & & & & \\
        \end{tabular}
        \caption{State 2}
    \end{subfigure}
    \begin{subfigure}[b]{0.3\textwidth}
        \begin{tabular}{ S | S | S | S | S }
            & & & & \\
            \hline
            & \emptycell & &\\
            \hline
            & & \cell{0.0} & \cell{0.0} & \\
            \hline
            & \cell{0.0} & & &  \\
            \hline
            & & \cell{0.0}& \cell{0.0} & \\
            \hline
            & & & & \\
        \end{tabular}
        \caption{State 3}
    \end{subfigure}
    \caption{Example of the Game of Life rules applied to a start state.}
    \label{fig:gol_example}
\end{figure}

To play Conway's Game of Life \cite{gol}, a player creates a state of living cells in a grid, after which they can observe the life inside evolve as defined by a set of rules.
This set consists of two rules, which both depend on the exact number of living neighbours.
The neighbourhood of a cell are those eight cells that directly surround it.
The rules of Life are as follows:
\begin{enumerate}
    \item A living cell survives to the next generation if it has exactly two or three living neighbours.
    \item A cell is born if it was dead in the previous generation, and had exactly three living neighbours.
\end{enumerate}
In this way, the first rule specifies the ``survival'' criterion for a living cell, and the second rule specifies the ``birth'' criterion for a dead cell to be born.
An example of these rules in action is shown in Figure \ref{fig:gol_example}.

Many extensions and variants exist for Game of Life, which can be categorised based on how they differ from the original.
The most straightforward variants simply alter the ruleset, such as the cellular automatons ``Flock'', in which cells can only survive with 1 or 2 neighbours, and ``Day and Night'', which has four rules for survival, and four rules for being born.
Other variants introduce more fundamental changes, such as converting Life into 3D \cite{Bays1991}, changing the size of the neighbourhood (e.g., Larger than Life \cite{Evans1996}), changing the grid to be non-square \cite{Bays2010}, and replacing the rules of Life by a neural network that learned to ``regrow'' certain patterns \cite{Mordvintsev2020}.

For this paper however, we focus on  those variants which add probability to the Game of Life -- in any form whatsoever.
For example, in \cite{Aguilera2019} the authors describe their Life extension ``Probabilistic Cellular Automata, Extension of the Game Of Life'' (PCAEGOL), in which the value of a neighbouring cell can be ``erroneous''.
Here, there is a certain probability of such an error occurring for each of the eight neighbours: for example, there could be a 20\% chance to count the left neighbour as being dead, while it actually is alive.
These errors are not consistent, in the sense that if a neighbour is considered erroneous for one cell, it might not be so for another cell.
The added probability of errors leads to the game becoming nondeterministic, where the same initial state can lead to different outcomes.

In \cite{Monetti1997}, the authors present Stochastic Game of Life (SGL).
This variant adds probability in two aspects of the game: (1) the survival rules have a probability $p_s$ associated to them, and (2) cells are always born if they have precisely three neighbours in the previous state, but also have a probability of $p_b$ to be born if they have precisely two neighbours.

Some works, like that of \cite{Schulman1978}, introduce a new stochastic component known as ``temperature'' $T$, which influences the probability of the rules in function of the \textit{density} of the grid.
Even further, regardless of the rules set by the player, $T$ can influence the life or death of a cell, acting as a way of introducing chaos into the system.

In Life, and most other cellular automata, the value of all cells is updated \textit{synchronously}, i.e., at the same time.
\cite{blok1999} presents an asynchronous variant, where each cell is no longer guaranteed to be updated at each time step, but instead only has a chance to do so.
Similar to the previous variants, this leads to a nondeterministic automaton.


Besides variants that introduce probability, there are also those variants that introduce continuous elements.
For example, SmoothLife, as introduced in \cite{Rafler2011}, transforms the rules of Life to work in a continuous grid, with a continuous function for the neighbourhood of a cell.
In \cite{Adachi2004}, the authors extend Life with continuous cell values, similar to this work.
They model the game's rules using a continuous transition function, which also contains a temperature component $T$.
As $T$ rises, the transition function becomes less precise, which in turn causes the cell values to become increasingly fuzzy, representing ``errors'' in the system.

\section{ProbLife}\label{s:problife}

In ProbLife, the value of a cell is no longer restricted to 0 (dead) or 1 (alive).
Instead, it can have any value of the continuous domain [0..1], where the value of a cell at time $t$ represents the probability that the cell is alive at that time.
For example, a cell value of 0.8 implies an 80\% chance of living.
Note that this preserves the meaning of 0 and 1 as guaranteed dead (0\% chance to live) and guaranteed alive (100\% chance to live).
The value of a cell in ProbLife is defined by a set of rules that denote the probability of a cell surviving or being born, given its exact number of living neighbours.
Such a rule, with a probability $x$ and a number of living neighbours $n$ is written as follows:
\begin{equation}
    p_{c}(n) = x.
\end{equation}
with $n$ an integer between 0 and 8, $c$ either ``$s$'' (survive) or ``$b$'' (birth), and $x$ a real number between 0 and 1.
For example, a rule stating that there is an 80\% chance for survival with exactly 4 neighbours is denoted by the following rule.

\begin{equation}
    p_{s}(4) = 0.8.
\end{equation}

The value of a cell at column $i$, row $j$ and time $t+1$ is then defined as:

\begin{equation}
    C_{t+1}(i, j) = \sum_{n=0}^{8} N_t(i,j,n) \times \bigg(p_s(n)\times C_{t}(i,j) + p_{b}(n)\times \big(1 - C_{t}(i,j)\big)\bigg).
\end{equation}
with $N_t(i, j, n)$ the probability that the cell at $(i,j)$ had $n$ neighbours alive at time $t$.

It is easy to see that ProbLife generalizes the original Game of Life, since we can recover the latter by the following ruleset:

\begin{equation}
    \begin{aligned}
        p_{b}(3) = 1.& \\
        p_{s}(2) = 1.& \\
        p_{s}(3) = 1.&\\
        p_e(i) = 0,& \text{ for all other } e \in \{s,c\} \text{ and } 0 \leq i \leq 8
    \end{aligned}
    \label{eq:problife_ruleset}
\end{equation}



\begin{figure}
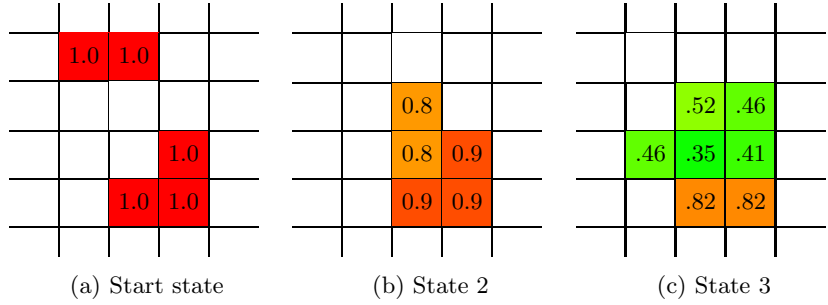

    \centering
    \begin{subfigure}[b]{0.3\textwidth}
        \begin{tabular}{ S | S | S | S | S }
            & & & & \\
            \hline
            & \probcell{0.0}{1.0} & \probcell{0.0}{1.0} & &\\
            \hline
            & & \emptycell & & \\
            \hline
            & & & \probcell{0.0}{1.0} &  \\
            \hline
            & & \probcell{0.0}{1.0} & \probcell{0.0}{1.0} & \\
            \hline
            & & & & \\
        \end{tabular}
        \caption{Start state}
    \end{subfigure}
    \begin{subfigure}[b]{0.3\textwidth}
        \begin{tabular}{ S | S | S | S | S }
            & & & & \\
            \hline
            & & \emptycell & &\\
            \hline
            & & \probcell{0.1}{0.8} & & \\
            \hline
            & & \probcell{0.1}{0.8} & \probcell{0.05}{0.9} &  \\
            \hline
            & & \probcell{0.05}{0.9} & \probcell{0.05}{0.9} & \\
            \hline
            & & & & \\
        \end{tabular}
        \caption{State 2}
    \end{subfigure}
    \begin{subfigure}[b]{0.3\textwidth}
        \begin{tabular}{ S | S | S | S | S }
            & & & & \\
            \hline
            & \emptycell & &\\
            \hline
            & & \probcell{0.24}{.52} & \probcell{0.27}{.46} & \\
            \hline
            & \probcell{0.27}{.46} & \probcell{0.325}{.35} & \probcell{0.295}{.41} &  \\
            \hline
            & & \probcell{0.09}{.82}& \probcell{0.09}{.82} & \\
            \hline
            & & & & \\
        \end{tabular}
        \caption{State 3}
    \end{subfigure}
    \caption{Example of ProbLife with $p_b(3) = 0.8$, $p_s(2) = 0.9$ and $p_s(3) = 0.9$.}
    \label{fig:example}
\end{figure}

Figure \ref{fig:example} shows an example of ProbLife in action.
Here, the probability for a cell to live is shown in two ways: (1) the colour of the cell, where red represents a high probability, blue represents a low probability and green represents a probability in between, and (2) the number in the cell, which corresponds directly to its chance to live.
The example uses a modified version of the standard Life ruleset, where the survival and birth probabilities have been set as respectively 90\% and 80\%: $p_s(2) = 0.9$, $p_s(3) = 0.9$, $p_b(3) = 0.8$.
We have found this ruleset to give good results, and will continue to use it for all other examples in this work as well.

Due to the probabilistic nature of ProbLife, cell configurations often die out completely after a few generations.
Indeed, on average the cell values will decrease with every generation, until the grid is empty.
While there is no way to reverse the decline, there are two main ways to get around this inevitable ``extinction'' by reaching stabilization.
The straightforward solution is to add rules with a probability of 1 in such a way that the cells stabilize after a few generations.
Alternatively, it is also possible to add a rule which causes dead cells with exactly 0 living neighbours to become alive (e.g. $p_b(0) = 0.8$), thereby turning ProbLife into a so-called ``strobing rule\footnote{\url{https://conwaylife.com/wiki/Strobing_rule}}''.
In this latter case, the grid can never be empty for more than one generation, i.e., it is not possible that every cell has a zero probability of being alive for two consecutive states.

\section{ProbLife in ProbLog}\label{s:problog}

This section shows that ProbLife can be elegantly implemented in the ProbLog~\cite{problog} system, a probabilistic extension of Prolog.
This allows for quick experimentation with different rulesets as a way to easily create prototypes.

A ProbLog program consists of a set of probabilistic facts, and a set of rules.
A probabilistic fact ``$P_f :: f$'' denotes a $P_f \in [0..1]$ probability for the atom $f$ to be true.
Rules in ProbLog are similar to those in Prolog, but with the addition of probabilities.
Concretely, they are of the form
\begin{equation}
    P_r::h \turnstile b_1,\ldots,b_n
\end{equation}
where the \textit{head} $h$ evaluates as true with a probability of $P_r$ if the \textit{body} $b_1,\ldots,b_n$ evaluates as true.
Here, the body consists of multiple \textit{body atoms} $b_i$, which all need to be true for the body to be true.
More information on the syntax and semantics of ProbLog can be found in \cite{problog}.
We can now translate the rules of ProbLife to ProbLog as follows.
A ProbLife rule $p_s(n) = z$ becomes:
\begin{equation*}
    \begin{aligned}
        z::\mathit{alive}(X,Y,T) \turnstile T > 0, T_p~is~T-1, \mathit{alive}(X,Y,T_p), \mathit{neigh}(X,Y,T_p,N).\\
    \end{aligned}
\end{equation*}
and a rule $p_b(n) = z$ is translated to:
\begin{equation*}
    \begin{aligned}
        z::\mathit{alive}(X,Y,T) \turnstile T > 0, T_p~is~T-1, not(\mathit{alive}(X,Y,T_p)), \mathit{neigh}(X,Y,T_p,N).
    \end{aligned}
\end{equation*}
where $\mathit{alive}(X,Y,T)$ is a predicate to represent a cell being alive at position $(X,Y)$ at time $T$, and $\mathit{neigh}(X,Y,T,N)$ a predicate to represent that the cell at $(X, Y)$ has exactly $N$ living neighbours at time $T$.
To denote the set of initially alive cells, we also introduce a predicate \textit{initAlive}$(X,Y)$, and add a rule that a cell is always alive at time $0$ if it is part of the initial cells.
For example, the ProbLife ruleset shown in (\ref{eq:problife_ruleset}) is represented in ProbLog as follows:
\begin{equation*}
    \begin{aligned}
        \mathit{alive}(X,Y,0) \turnstile \mathit{initAlive}(X,Y).\\
        0.9::\mathit{alive}(X,Y,T) \turnstile T > 0, T_p~is~T-1, not(\mathit{alive}(X,Y,T_p)), \mathit{neigh}(X,Y,T_p,3).\\
        0.9::\mathit{alive}(X,Y,T) \turnstile T > 0, T_p~is~T-1, not(\mathit{alive}(X,Y,T_p)), \mathit{neigh}(X,Y,T_p,2).\\
        0.8::\mathit{alive}(X,Y,T) \turnstile T > 0, T_p~is~T-1, \mathit{alive}(X,Y,T_p), \mathit{neigh}(X,Y,T_p,3).
    \end{aligned}
    \label{eq:problog_example}
\end{equation*}

Finally, the grid size and the initial state of a ProbLife instance are defined via ProbLog facts.
E.g., the example shown in Fig. \ref{fig:example} contains the following facts:
\begin{equation*}
    \small
    \begin{aligned}
        \mathit{row}&(0). & \mathit{row}&(1). &  \mathit{row}&(2). & \mathit{row}&(3). & \mathit{row}&(4). &\\
        \mathit{col}&(0). & \mathit{col}&(1). &  \mathit{col}&(2). & \mathit{col}&(3). & \mathit{col}&(4). &\\
        \mathit{initAlive}&(1,1). & \mathit{initAlive}&(2,1). & \mathit{initAlive}&(3,3). & \mathit{initAlive}&(2,4). & \mathit{initAlive}&(3,4) &
    \end{aligned}
\end{equation*}

A full ProbLog implementation of ProbLife is available online\footnote{\url{https://gitlab.com/EAVISE/sva/ProbLife}}, together with the code for a ProbLife editor.


\section{Discussion \& Conclusion}\label{s:discussion}

We will now briefly compare ProbLife to the other probabilistic variants described in Section \ref{s:gol}.
The main difference with the other automatons is that, while ProbLife introduces probability, the process of generating new states is actually deterministic.
Indeed, instead of making a choice between setting a cell as alive or dead, we instead assign it the probability of being alive.
When re-running the same initial state, these living probabilities will always remain the same.
Using ProbLog it is also possible to generate a sample run of a ProbLife system, i.e., one of the possible worlds.
In this case the cells would be binary again, instead of continuous, and would represent one possible outcome.
E.g., the Life example shown in Figure \ref{fig:gol_example} is a sample of the ProbLife example shown in Figure \ref{fig:example}.
Moreover, it is possible to use ProbLife together with ProbLog's sampling feature to recreate the probabilistic automaton presented in~\cite{Monetti1997}.

A possible application of ProbLife is the modelling of uncertainty.
E.g., consider a state in which there is a cell for which we are only 80\% sure that it is alive.
Using ProbLife, we could model the effect of this uncertainty on the rest of the population.
Another application is the propagation of an infection on a population, using a modified version of ProbLife in which cells can be designated as sick, healthy and/or vaccinated.
Using a specific ruleset (e.g., 80\% chance of evolving in a sick cell if unvaccinated, but only 5\% if vaccinated), we could model infections between generations.

\begin{figure}
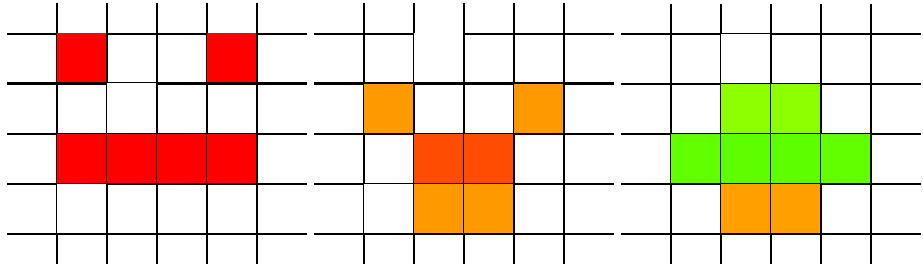

    \begin{tabular}{ S | S | S | S | S | S }
        &  & & & &\\
        \hline
        & \cell{0.0} & & & \cell{0.0} &\\
        \hline
        & & \emptycell & & & \\
        \hline
        & \cell{0.0} & \cell{0.0} & \cell{0.0} & \cell{0.0}&\\
        \hline
        & \emptycell & & & &  \\
        \hline
        & & & & &  \\
    \end{tabular}
    \begin{tabular}{ S | S | S | S | S | S }
        &  & & & &\\
        \hline
        & & \emptycell & & & \\
        \hline
        & \cell{0.1} & & & \cell{0.1} &\\
        \hline
        & & \cell{0.05} & \cell{0.05} & &\\
        \hline
        & \emptycell & \cell{0.1}& \cell{0.1}& &  \\
        \hline
        & & & & &  \\
    \end{tabular}
    \begin{tabular}{ S | S | S | S | S | S }
        &  & & & &\\
        \hline
        & & \emptycell & & & \\
        \hline
        & & \cell{0.2408} & \cell{0.2408}& &\\
        \hline
        & \cell{0.27} & \cell{0.27288} & \cell{0.27288} & \cell{0.27}&\\
        \hline
        & & \cell{0.1049} & \cell{0.1049}& &  \\
        \hline
        &  & & & &  \\
    \end{tabular}
    \caption{``Unamused tree''}
    \label{fig:art1}
\end{figure}

\begin{figure}
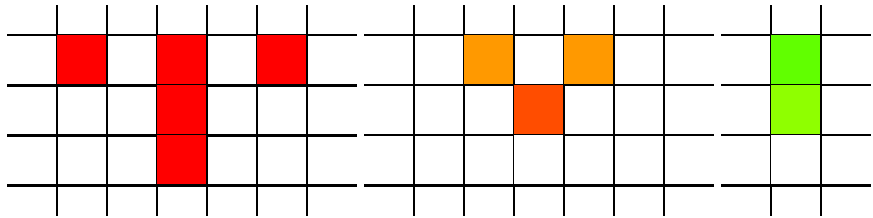

    \begin{tabular}{ S | S | S | S | S | S  | S}
        &  & & & & &\\
        \hline
        & \cell{0.0} & & \cell{0.0} & & \cell{0.0} & \\
        \hline
        & &  & \cell{0.0} & & & \\
        \hline
        &  & & \cell{0.0} & & &\\
        \hline
        & & & & & & \\
    \end{tabular}
    \begin{tabular}{ S | S | S | S | S | S  | S}
        &  & & & & &\\
        \hline
        & & \cell{0.1} & & \cell{0.1} & & \\
        \hline
        & &  & \cell{0.05} & & & \\
        \hline
        &  & & \emptycell & & &\\
        \hline
        & & & & & & \\
    \end{tabular}
    \begin{tabular}{   S | S | S  }
          & & \\
        \hline
        & \cell{0.2696}& \\
        \hline
        & \cell{0.24} & \\
        \hline
        & \emptycell & \\
        \hline
        & & \\
    \end{tabular}
    \caption{``Reverse Butterfly'', or, ``Cold Water''}
    \label{fig:art2}
\end{figure}

\begin{figure}
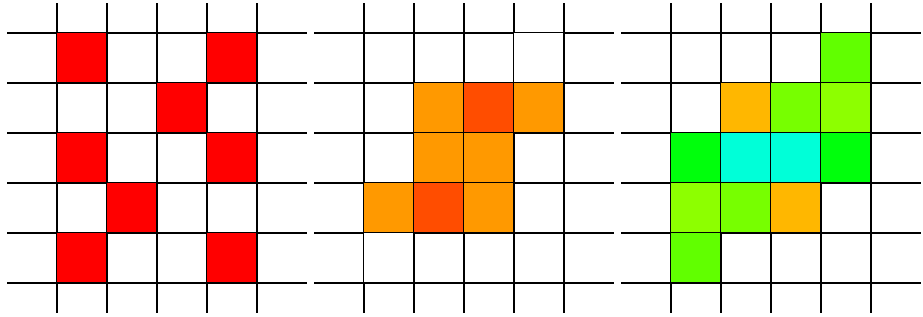

    \begin{tabular}{  S | S | S | S | S | S }
        & & & & & \\
        \hline
        & \cell{0.0} & & & \cell{0.0} & \\
        \hline
        & &  &  \cell{0.0} & &\\
        \hline
        & \cell{0.0} &  & & \cell{0.0} &  \\
        \hline
        &  & \cell{0.0} & & &\\
        \hline
        &  \cell{0.0} &  & & \cell{0.0} &\\
        \hline
         & & & & & \\
    \end{tabular}
    \begin{tabular}{  S | S | S | S | S | S }
        & & & & & \\
        \hline
        &  & & & \emptycell & \\
        \hline
        & & \cell{0.1} & \cell{0.05} & \cell{0.1} & \\
        \hline
        &  & \cell{0.1} & \cell{0.1} &  &\\
        \hline
        &  \cell{0.1} & \cell{0.05} & \cell{0.1} & &\\
        \hline
        &  \emptycell &  & &  &\\
        \hline
         & & & & & \\
    \end{tabular}
    \begin{tabular}{  S | S | S | S | S | S }
        & & & & & \\
        \hline
        & & & & \cell{0.2696} & \\
        \hline
        & & \cell{0.11984} &  \cell{0.2555} & \cell{0.2408} &\\
        \hline
        & \cell{0.34128} & \cell{0.4742305} & \cell{0.4742305} & \cell{0.34128} &  \\
        \hline
        & \cell{0.2408} & \cell{0.2555} & \cell{0.11984} & &\\
        \hline
        & \cell{0.2696} &  & &  & \\
        \hline
        & & & & & \\
    \end{tabular}
    \caption{``Fata Morgana'' starts as heat lines and ends in a lush, green oasis}
    \label{fig:art3}
\end{figure}

As mentioned earlier, generating images based on an initial state and a set of rules can be seen as a form of rule-based generative art.
Using our ProbLog implementation, we experimented with many different initial states and rulesets in order to look for any interesting formations.
The three most interesting ones of these are shown in Figures \ref{fig:art1}, \ref{fig:art2} and \ref{fig:art3}.


To conclude, this paper presents a probabilistic extension of Game of Life, called ProbLife.
It distinguishes itself in the fact that its cells have continuous cell values in the range of $[0..1]$, and that it remains deterministic.
Each rule in ProbLife has a probability associated to it, meaning that it is possible for a rule not to be applied.
Instead of the state of a cell being limited to either dead or alive, a cell in ProbLife is represented by its chance to live.
We modelled a concrete implementation of ProbLife in ProbLog, as a way to straightforwardly experiment with different rulesets and initial states.

\bibliography{biblio}{} 
\bibliographystyle{splncs04}

\end{document}